\documentclass[12pt]{article}
\usepackage{setspace}
\usepackage[utf8]{inputenc}
\usepackage{amsmath}
\usepackage{lineno}
\usepackage{esint}
\usepackage[hidelinks]{hyperref}
\usepackage{graphicx}
\usepackage[english]{babel}
\usepackage[nottoc]{tocbibind}
\usepackage{chngcntr}
\usepackage{array}
\newcolumntype{P}[1]{>{\centering\arraybackslash}p{#1}}
\usepackage{float}
\usepackage{subcaption}
\usepackage{algorithm}
\usepackage{forloop}
\usepackage{amsfonts}
\usepackage{algpseudocode}
\usepackage{subcaption}

\usepackage[
backend=biber,
style=numeric,
sorting=none
]{biblatex}
\graphicspath{ {./images/} }
\usepackage{geometry}
\usepackage{authblk}

\title{Continuous football player tracking from discrete broadcast data}
\author[1]{Matthew J. Penn}
\author[1,2]{Christl A. Donnelly}
\author[3,4]{Samir Bhatt}
\affil[1]{Department of Statistics, University of Oxford, Oxford, United Kingdom}
\affil[2]{Pandemic Sciences Institute, University of Oxford, Oxford, United Kingdom}
\affil[3]{Section of Epidemiology, University of Copenhagen, Copenhagen, Denmark}
\affil[4]{MRC Centre for Global Infectious Disease Analysis, Imperial College London, London, United Kingdom}
\begin{document}
\maketitle
\begin{abstract}
\noindent Player tracking data remains out of reach for many professional football teams as their video feeds are not sufficiently high quality for computer vision technologies to be used. To help bridge this gap, we present a method that can estimate continuous full-pitch tracking data from discrete data made from broadcast footage. Such data could be collected by clubs or players at a similar cost to event data, which is widely available down to semi-professional level. We test our method using open-source tracking data, and include a version that can be applied to a large set of over 200 games with such discrete data.
\end{abstract}

\subsection*{Introduction}
Data analysis is playing an increasingly important role for a growing number of football clubs at a wide range of levels. A key driver underpinning this growth is the availability of match event data - a set of events which describe the on-ball actions in a football match. Because it can be collected from broadcast footage through manual processing, event data is available for an increasingly large number of divisions, allowing a wide range of professional and semi-professional clubs to invest in obtaining and understanding the insights it provides. Crucial to this availability is the fact that the amount of event data per match is sufficiently small for it to be feasibly produced from match video by humans \cite{vidal2022automatic} - something which is necessary at lower levels of football as the quality of the footage is generally too low to enable computer vision algorithms to automate the process.
\\
\\
\noindent
However, the usefulness of event data is limited by the lack of context it provides on each action. The gold standard of match data is player tracking data \cite{barris2008review}, where the position of each player (and the ball) is provided at high frame rates, and access to this remains limited. Producing this data by hand would be incredibly time-consuming and many clubs will not record the whole pitch but will instead capture only the area close to the ball, meaning that a number of players are not visible, so even manual input would lead to partially incomplete datasets. Moreover, open-source player tracking data is extremely sparse, with only around 20 available matches, making it difficult for researchers and smaller football clubs to train and develop models without purchasing expensive datasets.
\\
\\
\noindent
To seek to remedy this problem, we introduce a novel method for estimating continuous, full-match tracking data from highly discrete, incomplete player positions, where only the team, but not the name of each player is recorded in each frame. We believe that creating such a dataset would be possible in a similar amount of time to event data, and industry examples already exist, such as Statsbomb's 360 data \cite{Statsbomb}. By providing code that can process Statsbomb 360 matches, we also enable researchers to access approximate continuous tracking datasets for around 200 matches \cite{Penn}. 
\\
\\
\noindent
There is a range of previous work which focuses on the creation of sports player tracking data. The focus in these studies is wide, including those which  focus on getting player trajectories directly from video footage \cite{xing2010multiple,thinh2019video,zhang2020multi}, predicting player trajectories using their past positions \cite{yeh2019diverse,hauri2021multi} or making predictions using players' posture \cite{suda2019prediction}. More closely related to our work, \cite{omidshafiei2022multiagent} examines a similar problem to the one we consider, attempting to predict player trajectories from broadcast data. However, these predictions were only made during short time intervals (9.5s) and using high-frequency input data (at 6.25 frames per second) and so therefore would not be applicable to our problem. 
\\
\\
\noindent
The authors of \cite{everett2023inferring} seek to address a similar issue to this paper by building a model that predicts all players' positions using only event data. This provides only one location per data point (equating to each player's position being recorded, on average, around once every 30 seconds), but provides scope for some positional analysis to be carried out on any current set of event data. However, as we will show later, our method provides a middle ground between full tracking data and this event-data-based approximation, with the extra data enabling us to achieve better results than \cite{everett2023inferring}, even when we compare our off-camera predictions to their overall error.
\\
\\
\noindent
The natural approach to this problem (used in, for example, \cite{omidshafiei2022multiagent} and \cite{everett2023inferring}) would be to make our prediction model using a Graph Neural Network. However, because of the sparsity of open-source training data (we use two matches compared to the 34 in \cite{everett2023inferring} and 105 in \cite{omidshafiei2022multiagent}, which were acquired directly from football data providers), and our aim to make a completely open-source model, we instead use a more parsimonious model, building a comparatively simple model based on autoregressive methods. This still provides a robust and useful prediction model, but we expect that it could be substantially improved upon with access to more data.
\subsection*{Methods}
\subsubsection*{Summary}
The overall method is as follows. Firstly, we will build a predictive model for a player's position over time based on their past positions and the ball position. Using this model, we will create player trajectories by assigning each of the set of observed player positions in each frame to a trajectory. Once this has been done, we will then build a model to interpolate between the points in each trajectory to create continuous player paths.
\\
\\
\noindent
Note that the aim of this code is not to track individual player trajectories, but to simply provide the positions of the players on each team. While tracking individual trajectories would be preferable (and is possible on short timescales), events such as corners - where players move, off-camera, from a variety of locations to all be in a very small area - mean that it is impossible to reliably track individual players over the course of the match. This could be remedied by labelling each player whose position was recorded with their name, although this would require substantial additional effort from the creators of the data.
\subsubsection*{Notation and preliminaries}
We suppose that the frames are sampled at times $t_1,...,t_n$ and that the ball position, $\boldsymbol{b}(t_i)$ is known at each frame $i$. 
\\
\\
\noindent
We assume throughout that the team, and whether or not they are an outfielder or a goalkeeper, of each player in the frame has also been provided, but that the specific identity of a player is not given. Such classification can be done easily using the shirt colour of the players, and is provided in datasets such as the Statsbomb 360 data  \cite{Statsbomb}. Thus we can separately estimate the trajectories of the players on each team (though our final interpolation stage will involve the players from both teams). 
\\
\\
\noindent
Following the convention used in the Statsbomb data, we assume that the pitch measures 120m by 80m. This means that, when using the Metrica Sports player tracking data \cite{Metrica}, where coordinates are given as percentages (i.e. the centre spot is (0.5,0.5) and the goals are (0,0.5) and (1,0.5)), we scale them to ensure they are comparable.
\subsubsection*{Player position forecasting}
Suppose that we know the past trajectory $X = \bigg\{(\boldsymbol{x}(t_1),t_1),...,(\boldsymbol{x}(t_m),t_m)\bigg\}$ up to time $t_m$ of a given player, where $\boldsymbol{x}(t_i)$ is the position at time $t_i$. Note that we may not necessarily know the position of the player at every sampled frame.
\\
\\
\noindent
We aim to define functions $\mu(t,\boldsymbol{b},X)$ and $\sigma(t,\boldsymbol{b},X)$ such that our predicted trajectory $\boldsymbol{x}(t)$ for $t > t_m$ is
\begin{equation}
    \boldsymbol{x}(t) \sim \text{N}\bigg(\mu(t,\boldsymbol{b},X) ,\sigma(t,\boldsymbol{b},X)^2I\bigg)
\end{equation}
where $I$ is the 2x2 identity matrix. Note that we will use the position of the ball after time $t_m$, as well as the positions before it.
\\
\\
\noindent
We do this by building a discrete autoregressive moving average with exogenous inputs (ARMAX) model for player positions, using the ball position as an exogenous variable (as players will, in general, move in a similar direction to the ball). In the method presented in this section, we will assume that we use a temporal grid size of one second, but other sizes could be used. We use $\boldsymbol{x}^k$ to denote the (estimated) position at time $k$. For $t_i < k < t_{i+1}$, we can estimate the value of $\boldsymbol{x}^k$ using a simple linear interpolation
\begin{equation}
    \boldsymbol{x}^k = \boldsymbol{x}(t_i) + \frac{k-t_i}{t_{i+1} - t_i}\bigg(\boldsymbol{x}(t_{i+1}) - \boldsymbol{x}(t_i)\bigg).
\end{equation}
We train our ARMAX model using player trajectories taken from Match 2 of the Metrica Sports open dataset \cite{Metrica} (we reserve Match 1 for testing our model). With more data, it would be possible to train separate models for different positions (as, for example, a left back would show far different movement to a right winger) but given the relatively small size of the data, we simply append all outfield player trajectories together to train a single ARMAX model. 
\\
\\
\noindent
When we predict $\boldsymbol{x}^{k+n}$ from the points $\boldsymbol{x}^1,...,\boldsymbol{x}^k$, we use the ARMAX model only when $n > 1$. For $n=1$ we instead suppose that $\boldsymbol{x}^{k+1} = N(\boldsymbol{x}^k,s^2I)$ for some constant $s$. For players such as central defenders with smaller responses to the ball movement, this prevents inaccurate estimates causing incorrect trajectory assignment.

\subsubsection*{Trajectory assignment}
For each of the 10 outfield players on a given team, we aim to build trajectories of known points (i.e. of player positions that are recorded in a frame). We suppose that the trajectory $\mathcal{T}_i$ of player $i$ is given by 
\begin{equation}
    \mathcal{T}_i = \bigg\{(\boldsymbol{x}_i(t^i_1),t^i_1),...,(\boldsymbol{x}_i(t^i_{m_i}),t^i_{m_i})\bigg\}.
\end{equation}
At each frame, we append each of the $N$ visible player positions to one of these trajectories (and must ensure that no trajectory has multiple positions from the same frame appended to it, as a player cannot simulataneously be in more than one position). Using our predictive model, the position of player $i$ at the frame time $t$ is given by
\begin{equation}
   \boldsymbol{x}_i(t)  \sim \text{N}\bigg(\mu(t,\boldsymbol{b},\mathcal{T}_i) ,\sigma(t,\boldsymbol{b},\mathcal{T}_i)^2I\bigg)
\end{equation}
and we can therefore define a matrix $Q$ by
\begin{equation}
    Q_{ij} = \log\bigg(\mathbb{P}(\text{player $i$ is at visible position $j$ at time $t$})\bigg).
\end{equation}
The maximum likelihood assignment of players to points can then be found using the Hungarian algorithm on the matrix $-Q$ (as, treating $-Q_{ij}$ as the cost of assigning position $j$ to trajectory $i$, minimizing the cost is the same as maximising the sum of the log-likelihoods, and therefore the overall log-likelihood). 
\subsubsection*{Player position initialisation}
Calculating the matrix $Q_{ij}$ requires some previous information on the location of the players. Generally, the majority of outfield players are visible in the initial frame. Those that are not visible are assigned to plausible positions along the defensive line. This may lead to some errors early on in the fitting procedure, but we have found these to be generally insubstantial.
\subsubsection*{Continuous path interpolation}
Given the final processed trajectories $\mathcal{T}_i$, we then aim to create a continuous path from the points it contains. To do this, we first calculate the \textit{average known velocity} $\boldsymbol{v}_k$ at each time $k$. This is done by defining $S_k$ to be the set of all outfield players (on both teams) such that their trajectories contain points at times $k+1$ and $k$, and then setting $\boldsymbol{v}_k$ to be the average difference between their recorded points at these times. Note that if no trajectories have positions at both times $k+1$ and $k$, we set $\boldsymbol{v}_k = \boldsymbol{0}$.
\\
\\
\noindent
Using a linear interpolation of these velocities, $\boldsymbol{u}(t)$, we can then define a weighted average velocity
\begin{equation}
    \boldsymbol{w}(s,t) = \alpha \int_s^t \boldsymbol{u}(r)dr
\end{equation}
for some constant $\alpha$. We have found that setting $\alpha \approx 0.5$ leads to the best results, which we believe is because players near the ball will be moving with higher velocity than players further away. 
\\
\\
\noindent
Once the velocities have been calculated, the continuous player paths are estimated as follows. Suppose that a trajectory $\mathcal{T}_i$ contains positions at times $t_1$ and $t_2$ and contains no positions in the interval $(t_1,t_2)$. Then, for $t \in (t_1,t_2)$, we set the path
\begin{equation}
\label{eq:interp}
    \boldsymbol{x}_i(t) = \boldsymbol{x}_i(t_1) + \boldsymbol{w}(t_1,t) + \bigg(\frac{t-t_1}{t_2-t_1}\bigg)\bigg(\boldsymbol{x}_i(t_2) - \boldsymbol{x}_i(t_1) - \boldsymbol{w}(t_1,t_2)\bigg).
\end{equation}
There are two components to this formula. Firstly, there is the simple linear interpolation between the positions, which would be the estimate if the velocity were ignored.
\begin{equation}
    \boldsymbol{x}_i(t_1) + \bigg(\frac{t-t_1}{t_2-t_1}\bigg)\bigg(\boldsymbol{x}_i(t_2) - \boldsymbol{x}_i(t_1)\bigg)
\end{equation}
Secondly, there is the velocity correction
\begin{equation}
     \boldsymbol{w}(t_1,t)  - \bigg(\frac{t-t_1}{t_2-t_1}\bigg)\boldsymbol{w}(t_1,t_2).
\end{equation}
This is the difference between the true velocity integral up to time $t$ and the linear approximation to it based only on the total velocity integral in the interval $(t_1,t_2)$. This ensures that players follow more realistic paths as their velocities are related to those of other players. Indeed, the overall formula could be derived by considering a linear interpolation to the difference between $\boldsymbol{x}_i$ and the ``visible centre of mass'' (an object moving according to the visible velocity $\boldsymbol{u}$).
\subsubsection*{Runtime}
Using the pre-trained ARMAX model, the vast majority of the runtime for the code is performing the trajectory assignments, which runs at approximately 0.23s/frame on a standard laptop through Jupyter Lab (Python) on a single core with a clock speed of 2.6GHz. This means that a match with one frame sampled per second can be processed in 20-25 minutes, depending on the amount of stoppage time. Matches with an extra-time period of course take proportionally longer.
\subsection*{Results}
\subsubsection*{Data}
We test our model using Match 1 of the Metric Sports open dataset \cite{Metrica}. The first half and second half are tested separately (as the players swapping sides during the interval would lead to substantial errors). Moreover, the first and last 30 frames of each half are ignored (as this contains the players entering and leaving the pitch, where they move in an unusual way and are not near the ball - again, including these frames would simply introduce an unrealistic level of errors).
\\
\\
\noindent
To simulate the effect of broadcast data, we sample frames every second and record only players who are within 30 metres of the ball. This leads to an average of 10.60 out of the 20 outfield players being visible, which is substantially less than the Statsbomb data (where on average 14.35 of the 20 outfield players were visible, though this ranged substantially from 10.93 to 18.58 depending on the match). Thus, we expect the results of this test will be robust to the change of dataset.
\\
\\
\noindent
There is, on average, a smaller number of available frames in the Statsbomb data - there are frames approximately every 1.9 seconds (rather than a frame every second in our test). However, rather than being uniformly spread, these frames are concentrated on action points in the match (where the players are moving quickly), rather than, for example, in the periods of time waiting for set pieces to be taken. With the ball in play for around 60\% of football matches \cite{tojo2023effective}, this means that the frame rates when the ball is in play will be comparable.
\\
\\
\noindent
Ignoring, as explained previously, the last 30 frames from each half, the total number of frames sampled is 6390 and the total number of position predictions is 59821.
\subsubsection*{Error analysis}
\begin{figure}[H]
    \centering
    \includegraphics[width = 1\textwidth]{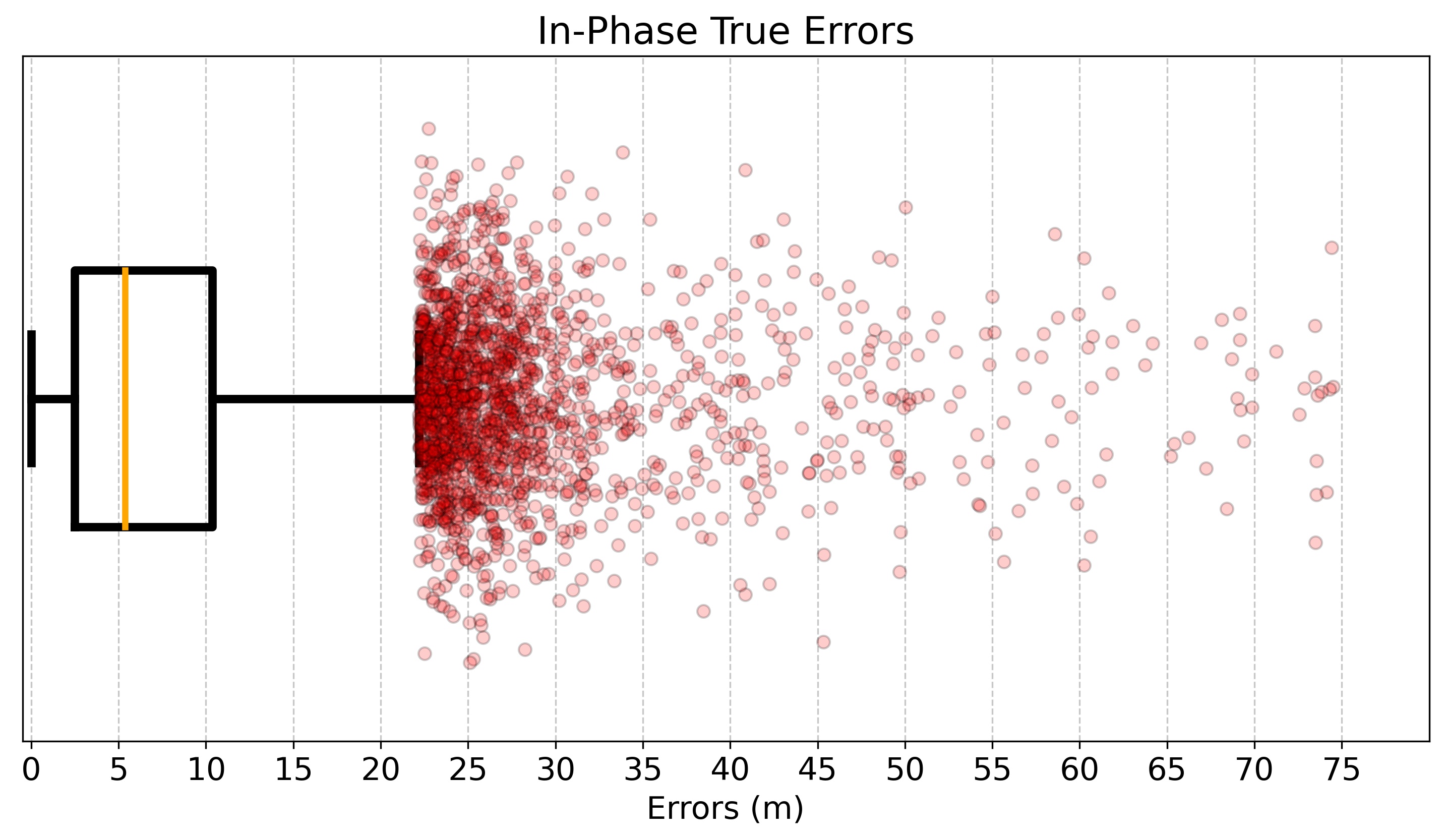}
    \caption{A boxplot showing the distribution of the in-phase true errors (that is, the errors for players not visible in the frame). The red dots show the distribution of errors above the 99th percentile.}
    \label{fig:Error_Boxplot}
\end{figure}

\begin{figure}[H]
    \centering
    \includegraphics[width = 1\textwidth]{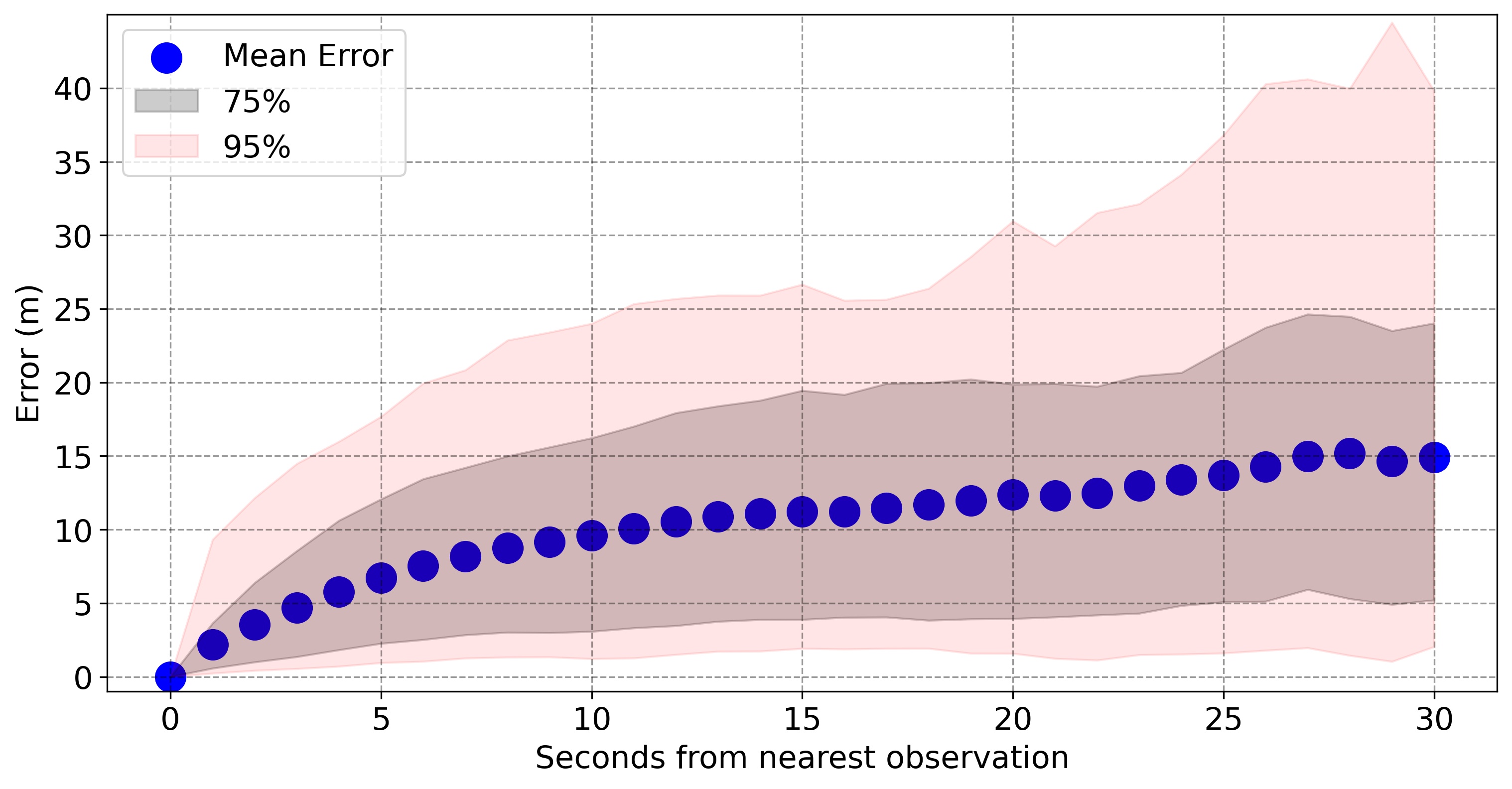}
    \caption{A plot showing the mean error and symmetric 75\% and 95\% intervals (shaded black and red regions) of the prediction errors depending on the amount of time between the prediction and the nearest observation (either forwards or backwards in time).}
    \label{fig:Error_Time}
\end{figure}

Recalling that we sample frames at a constant rate of one-per-second, we consider the errors in our estimation both ``in phase'' (i.e. after an integer number of seconds) and ``out of phase'' (i.e. after a half-integer number of seconds). We also consider the overall error (including the players for whom we know the position exactly) and for players that are ``off-camera'' - that is, the errors in estimating the positions of players with positions that were not recorded.
\\
\\
\noindent
To calculate the error at a given time, we first map the set of estimated player positions on each team to the set of true positions using the Hungarian Algorithm, where the cost is the distance between the estimated and true position. Our error for a given player is then the distance between its estimated position and the true position to which it has been assigned. 
\\
\\
\noindent
When we consider the times that are in phase, many of the ``estimates'' are exact, as, from the interpolation (\ref{eq:interp}), the continuous player path $\boldsymbol{x}_i(t)$ interpolates the player trajectory $\mathcal{T}_i$. Thus, while the mean error for all players is 3.45m, a more realistic metric is the mean error for those players out of the frame, which is 7.33m. This mean is dominated by some very high errors which can be caused, as evidenced by the fact that the median off-camera error is substantially lower at 5.37m. The distribution of these errors is shown in Figure \ref{fig:Error_Boxplot}. 
\noindent
The out-of-phase results are similar, with an overall mean error of 3.55m. Again, this is substantially reduced by players whose positions were recently observed, highlighted by the fact that the error for players whose position was observed in the previous frame is 0.22m.
\\
\\
\noindent
One of the major sources of error is the frames where the ball is ``dead'' (for example, if play has been paused because of a free kick or throw-in). This is highlighted by the fact that if only the in-phase frames closest to those in which an event has been recorded in the associated event data provided by Metrica Sports are used, the mean off-camera error drops dramatically to 6.62m.
\\
\\
\noindent
Unsurprisingly, the error is highly correlated with the amount of time between the prediction and the nearest observation (e.g. if a player's trajectory contained points at times 5s and 10s and the prediction was made at time 8s, this time to nearest observation would be 2s). Figure \ref{fig:Error_Time} shows that the error grows relatively slowly, with players off-camera for 20 seconds, and therefore with a maximum of 10 seconds to the nearest observation still having a mean error of under 10m and a $97.5^{\text{th}}$ percentile of less than 25m. Given that a professional athlete could easily run between any two points on the pitch in 20s (a maximum distance of 145m), this shows the ability of the model to provide sensible predictions even when there is very little data.
\subsubsection*{Examples}
To illustrate the performance of the model, after discarding the first and last 30 frames, we choose the first frame in the first half of Match 1 where the total squared error (that is, the total sum of the squared prediction errors at that frame) is closest to the $25^{\text{th}}$, $50^{\text{th}}$, $75^{\text{th}}$ and $95^{\text{th}}$ percentile of all frame errors. This provides a fair overview of the model performance, and puts these errors into the context of the match.
\begin{figure}[H]
    \begin{subfigure}{0.5\textwidth} 
        \includegraphics[width=\textwidth]{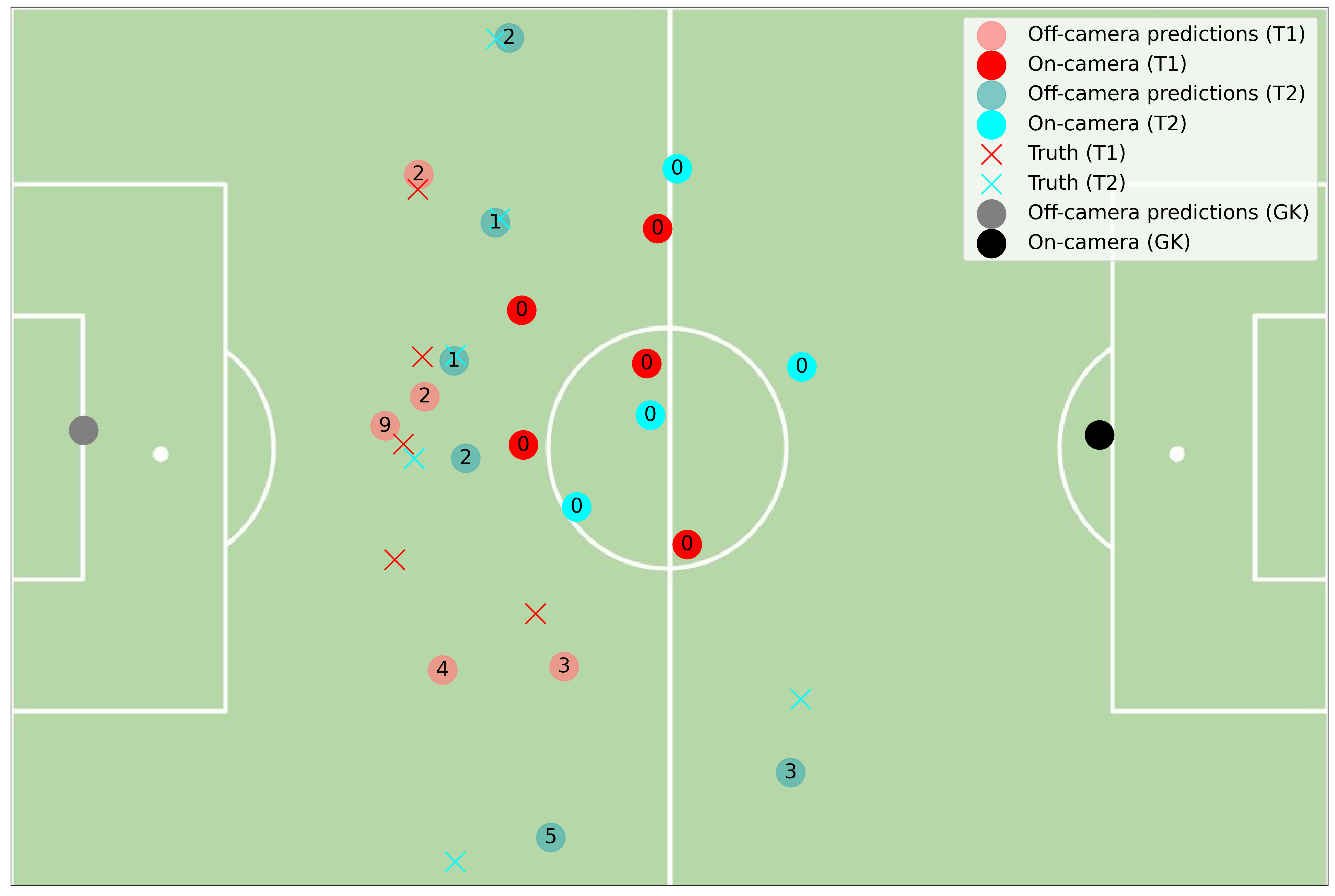}
        \caption{$25^{\text{th}}$ percentile}
        \label{subfig:a}
    \end{subfigure}
    \begin{subfigure}{0.5\textwidth} 
        \includegraphics[width=\textwidth]{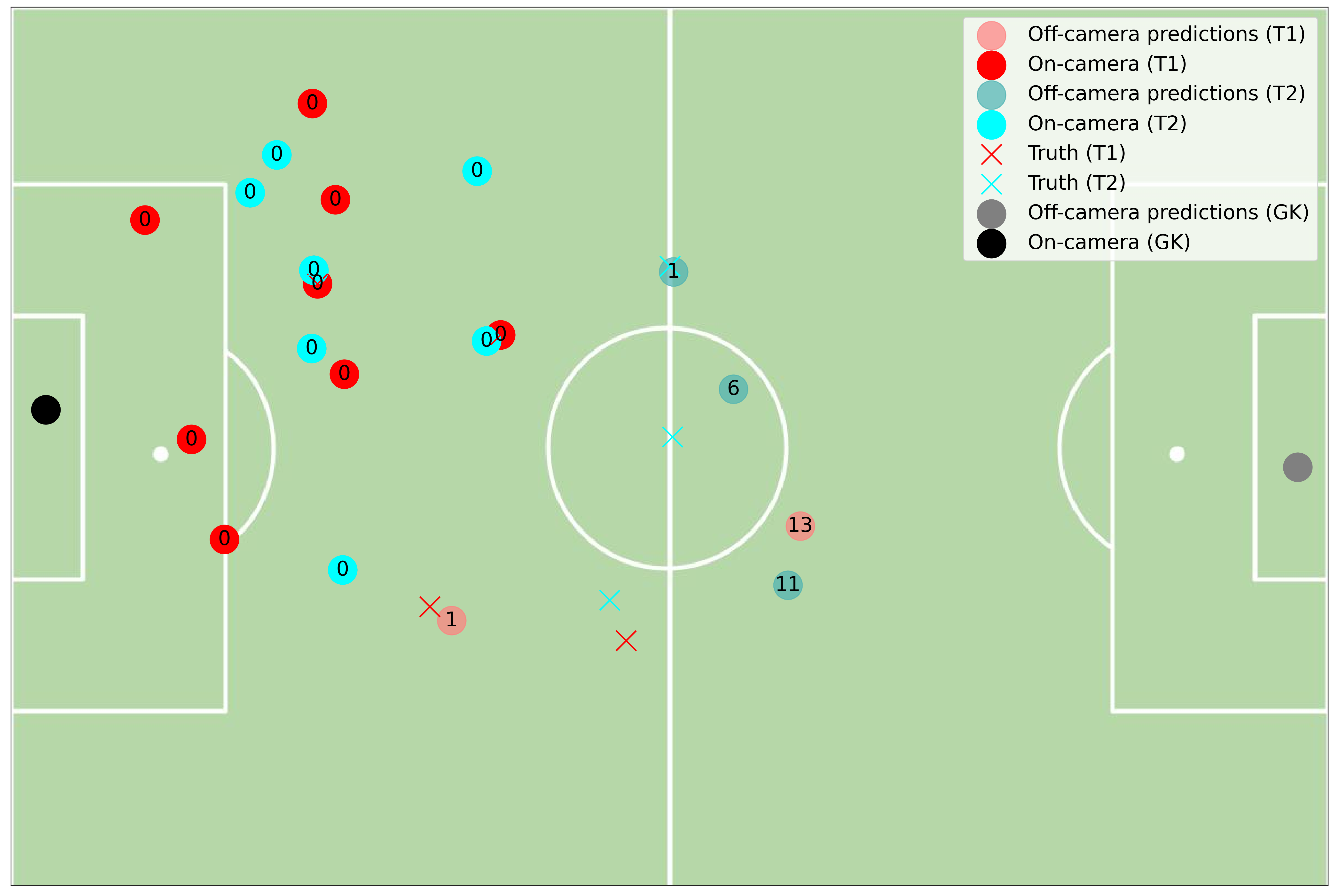}
        \caption{$50^{\text{th}}$ percentile}
        \label{subfig:b}
    \end{subfigure}
    
    \begin{subfigure}{0.5\textwidth} 
        \includegraphics[width=\textwidth]{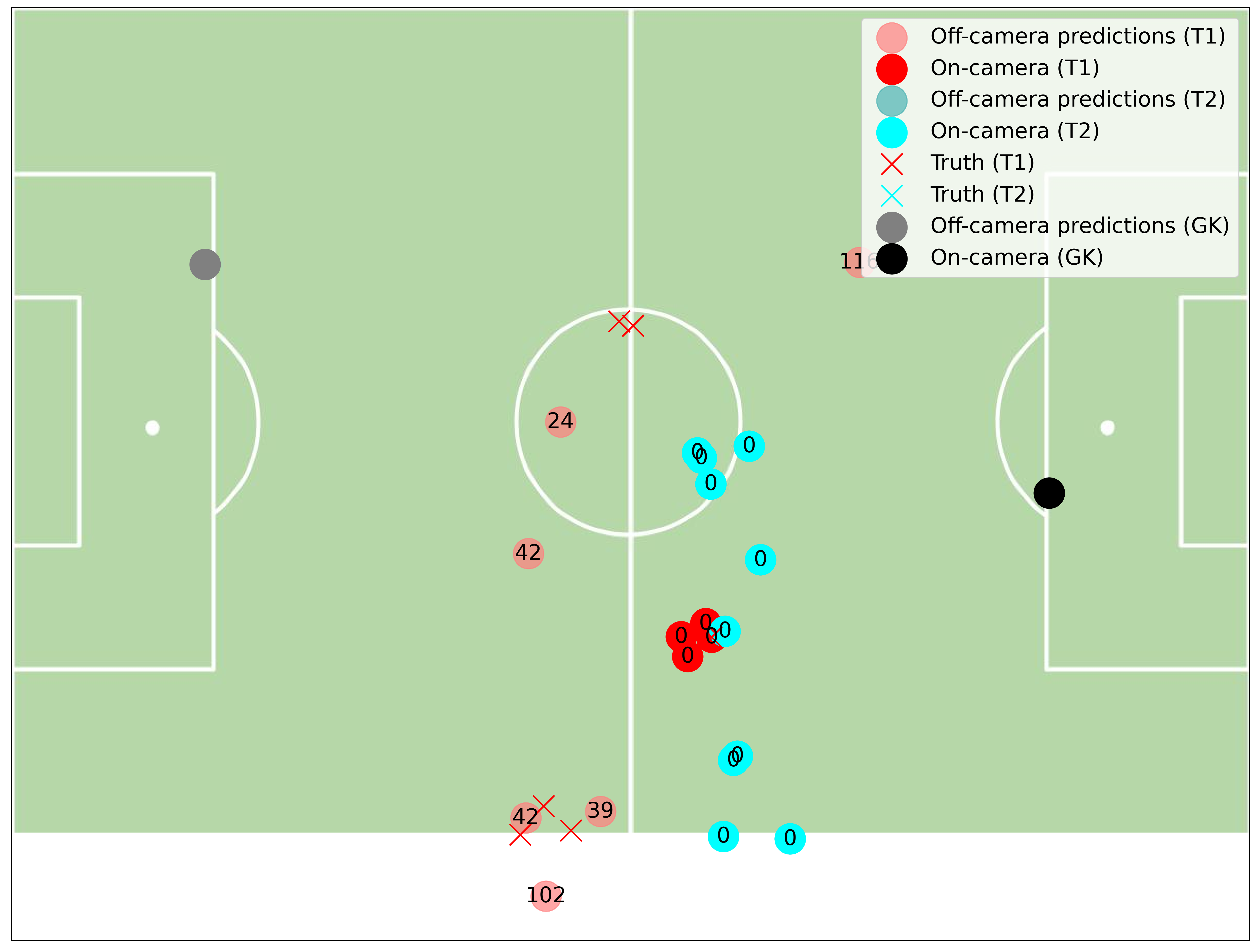}
        \caption{$75^{\text{th}}$ percentile}
        \label{subfig:c}
    \end{subfigure}
    \begin{subfigure}{0.5\textwidth} 
        \includegraphics[width=\textwidth]{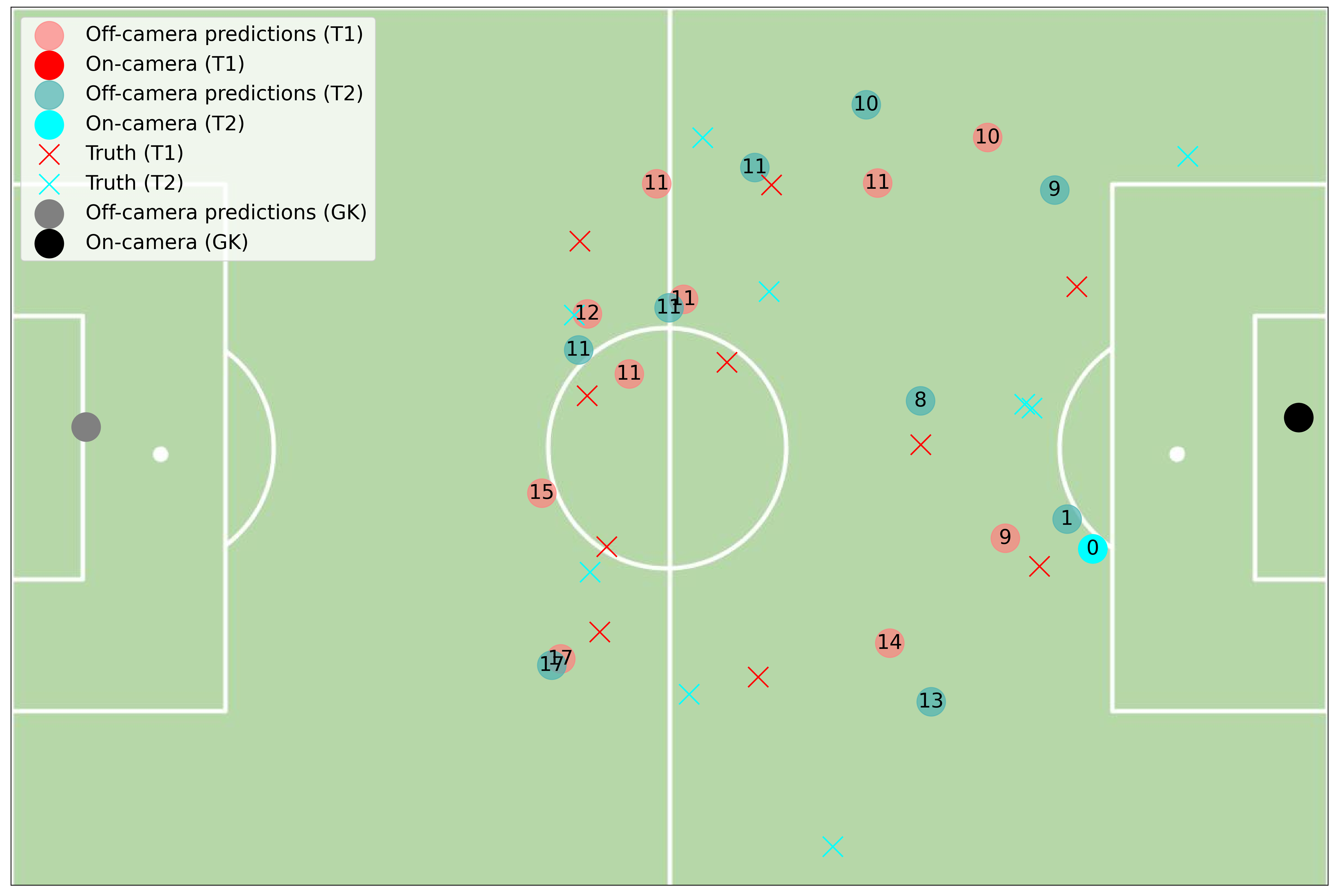}
        \caption{$95^{\text{th}}$ percentile}
        \label{subfig:d}
    \end{subfigure}
    
    \caption{An illustration of model performance based on frames where the total squared prediction error is at different percentiles. The numbers inside each of the dots are the number of frames to the closest frame in which that player was observed (according to the trajectories created by the model). Note that goalkeepers' predicted positions are included, but were not part of the error analysis. In all examples, the red team is attacking from left to right.}
    \label{fig:main}
\end{figure}
\noindent
These examples are shown in Figure \ref{fig:main}. At the $25^{\text{th}}$ percentile, Figure \ref{subfig:a} provides an example where, despite the majority of the players being out of frame the errors are small and, importantly, the overall structure of both teams remains the same. This includes one of the red central defenders (with a ``9'' in the middle of their circle) even though the closest frame in which they are visible is 9 frames (or, equivalently, seconds) away. Being able to maintain an accurate prediction of their position despite being off-camera for a total of at least 18 seconds highlights the ability of our algorithm to produce useful predictions.
\\
\\
\noindent
Figure \ref{subfig:b} is at the $50^{\text{th}}$ percentile of total error. In this frame, the majority of players are visible, with two players (the furthest right blue and red circles) having been off-camera for a long time - at least 22 seconds. Their predictions are reasonably inaccurate, but importantly, the structure of both teams is still roughly correct - the red player is playing as a single striker towards the right-hand side of the pitch, while the blue defender is marking him. Thus, using these positions to aid prediction of the goalscoring potential of each team is likely to have small errors (much less so than, for example, if the predictions for these players were very far apart from each other).
\\
\\
\noindent
Figure \ref{subfig:c} illustrates the source of a substantial proportion of the high-error frames. Using the associated event data provided by Metrica Sports, we can see that this frame is in the middle of a 3-minute-long stoppage following a foul. During this time, the players move in unusual patterns (with, for example, many players grouped closely together), while players on the other side of the pitch are off-camera for an extended period of time (the rightmost red player was off-camera for around 4 minutes). Knowing the positions of the players during this time is much less important as the ball is ``dead'' and players are slowly moving towards their positions for the freekick. However, again, the overall structure is mostly maintained, illustrating the robustness of our model.
\\
\\
\noindent
Finally, Figure \ref{subfig:d}, at the $95^{\text{th}}$ percentile of total error, is an example of the model performance when most players are off-camera for a large period of time. This occurs particularly regularly (as in this frame) at goal kicks. This leaves the model to be susceptible to error when the off-camera players move substantially while they are hidden. This was the case while the teams waited for this goal kick to be taken, as one of the pair of close-together blue players on the edge of their D (the curved region on the edge of the penalty box) moved to their right-hand side (that is, the top of the pitch as presented in the figure). Thus, the errors in team structure that are visible in this figure are much reduced by the time the goal-kick is taken - indeed, the total squared error reduces from 2950$\text{m}^2$ to 900$\text{m}^2$ in the frame before the goal-kick is taken (despite the fact that none of the outfield players are visible in this frame). Thus, the impact of these errors is minimal as they would have little impact on the evaluation of the game, and they do not cause any mis-assignment of player trajectories as none of the players are visible during this time.
\subsubsection*{Statsbomb Data}
We have also produced a version of the code that can process Statsbomb's discrete broadcast data (their ``360 data''). This code is available at \cite{Penn}. The processed data is given in the same JSON format as Statsbomb's data with an additional field to show whether the player was visible in the frame, or whether the position has been estimated by our algorithm. A very small percentage of the frames will return an error due to large disagreements between the position of the ball in the 360 data and the associated event data, also found at \cite{Statsbomb} (this event data was used as the Statsbomb data measures its x coordinates from the defending team's goal line, meaning that the axes are not fixed). These errors will be recorded in our outputted files.
\subsection*{Discussion}
The position prediction model introduced in this paper has the potential to enrich discrete, incomplete datasets of player positions created from broadcast data. It performed well on our test dataset, achieving a mean error of 7.33m for non-visible players. We believe this is sufficiently accurate to be useful for a wide range of modelling tasks, as in the context of a football pitch, it is a relatively small distance (for example, a circle of radius 7.33m would fill only 1.75\% of a football pitch).
\\
\\
\noindent
Our model compares well to the results presented in the event data model \cite{everett2023inferring}. Rescaling their pitch to be the same length as ours (their pitch measured 105m by 68m, so all distances are multiplied by $120/105$), their mean error was 7.86m. This is over double our mean error of 3.45m (of course, this is largely due to the fact that our model has more data), and higher than our off-camera means at events (6.62m) and at all frames (7.33m). While the difference between our off-camera prediction error and the overall error in \cite{everett2023inferring} is not as large as perhaps would be expected, players closer to the ball are much easier to predict (as a triviality, the player carrying out the event can be predicted with zero error in \cite{everett2023inferring} but would be excluded from our off-camera dataset) and so it seems likely that the mean error in the \cite{everett2023inferring} would increase substantially if only off-camera players were included. However, including some of the features of the model in \cite{everett2023inferring} could improve our prediction of player positions when they have been off-camera for a substantial amount of time.
\\
\\
\noindent
There are still limitations to our outputs. While our level of error would allow, for example, teams' attacking and defensive structures to be investigated, it would be difficult to use the estimated positions to examine attributes such as the positioning of off-camera players, where the impact of moving a couple of metres can be substantial. Use of this data would require a careful consideration of the intrinsic error, and any analysis that requires a high level of accuracy in the data has the potential to produce misleading results. However, we believe our algorithm provides a substantial improvement on event data and would be a key addition to clubs' datasets.
\maketitle
\printbibliography

@misc{Metrica,
  author = {metrica-sports},
  title = {Metrica Sports Sample Data},
  year = {2021},
  publisher = {GitHub},
  journal = {GitHub repository},
  howpublished = {\url{https://github.com/metrica-sports/sample-data}},
  commit = {e706dd5}
}

@misc{Penn,
  author = {Matthew J. Penn},
  title = {Football-Tracking-Interpolation},
  year = {2023},
  publisher = {GitHub},
  journal = {GitHub repository},
  howpublished = {\url{https://github.com/mpenn114/Football-Tracking-Interpolation}},
  commit = {5808d6c}
}

@misc{Statsbomb,
  author = {statsbomb},
  title = {open-data},
  year = {2023},
  publisher = {GitHub},
  journal = {GitHub repository},
  howpublished = {\url{https://github.com/statsbomb/open-data}},
  commit = {5401c04}
}

@article{tojo2023effective,
  title={Effective playing time affects technical-tactical and physical parameters in football},
  author={Tojo, {\'O}scar and Spyrou, Konstantinos and Teixeira, Jo{\~a}o and Pereira, Paulo and Brito, Jo{\~a}o},
  journal={Frontiers in Sports and Active Living},
  volume={5},
  year={2023},
  publisher={Frontiers Media SA}
}

@article{xing2010multiple,
  title={Multiple player tracking in sports video: A dual-mode two-way bayesian inference approach with progressive observation modeling},
  author={Xing, Junliang and Ai, Haizhou and Liu, Liwei and Lao, Shihong},
  journal={IEEE Transactions on Image Processing},
  volume={20},
  number={6},
  pages={1652--1667},
  year={2010},
  publisher={IEEE}
}

@inproceedings{thinh2019video,
  title={A video-based tracking system for football player analysis using efficient convolution operators},
  author={Thinh, Nguyen Hong and Son, Hoang Hong and Dzung, Chu Thi Phuong and Dzung, Vu Quang and Ha, Luu Manh},
  booktitle={2019 International Conference on Advanced Technologies for Communications (ATC)},
  pages={149--154},
  year={2019},
  organization={IEEE}
}

@article{zhang2020multi,
  title={Multi-camera multi-player tracking with deep player identification in sports video},
  author={Zhang, Ruiheng and Wu, Lingxiang and Yang, Yukun and Wu, Wanneng and Chen, Yueqiang and Xu, Min},
  journal={Pattern Recognition},
  volume={102},
  pages={107260},
  year={2020},
  publisher={Elsevier}
}

@inproceedings{yeh2019diverse,
  title={Diverse generation for multi-agent sports games},
  author={Yeh, Raymond A and Schwing, Alexander G and Huang, Jonathan and Murphy, Kevin},
  booktitle={Proceedings of the IEEE/CVF Conference on Computer Vision and Pattern Recognition},
  pages={4610--4619},
  year={2019}
}

@inproceedings{hauri2021multi,
  title={Multi-modal trajectory prediction of NBA players},
  author={Hauri, Sandro and Djuric, Nemanja and Radosavljevic, Vladan and Vucetic, Slobodan},
  booktitle={Proceedings of the IEEE/CVF Winter Conference on Applications of Computer Vision},
  pages={1640--1649},
  year={2021}
}

@inproceedings{suda2019prediction,
  title={Prediction of volleyball trajectory using skeletal motions of setter player},
  author={Suda, Shuya and Makino, Yasutoshi and Shinoda, Hiroyuki},
  booktitle={Proceedings of the 10th Augmented Human International Conference 2019},
  pages={1--8},
  year={2019}
}

@article{omidshafiei2022multiagent,
  title={Multiagent off-screen behavior prediction in football},
  author={Omidshafiei, Shayegan and Hennes, Daniel and Garnelo, Marta and Wang, Zhe and Recasens, Adria and Tarassov, Eugene and Yang, Yi and Elie, Romuald and Connor, Jerome T and Muller, Paul and others},
  journal={Scientific reports},
  volume={12},
  number={1},
  pages={8638},
  year={2022},
  publisher={Nature Publishing Group UK London}
}

@article{everett2023inferring,
  title={Inferring Player Location in Sports Matches: Multi-Agent Spatial Imputation from Limited Observations},
  author={Everett, Gregory and Beal, Ryan J and Matthews, Tim and Early, Joseph and Norman, Timothy J and Ramchurn, Sarvapali D},
  journal={arXiv preprint arXiv:2302.06569},
  year={2023}
}

@article{barris2008review,
  title={A review of vision-based motion analysis in sport},
  author={Barris, Sian and Button, Chris},
  journal={Sports Medicine},
  volume={38},
  pages={1025--1043},
  year={2008},
  publisher={Springer}
}

@article{vidal2022automatic,
  title={Automatic event detection in football using tracking data},
  author={Vidal-Codina, Ferran and Evans, Nicolas and El Fakir, Bahaeddine and Billingham, Johsan},
  journal={Sports Engineering},
  volume={25},
  number={1},
  pages={18},
  year={2022},
  publisher={Springer}
}

\end{document}